\documentclass[11pt,a4paper]{article}
\usepackage[hyperref]{emnlp2020}
\usepackage{times}
\usepackage{amsmath}
\usepackage{latexsym}
\usepackage{graphicx}
\usepackage{bbm}
\usepackage{stfloats}
\usepackage{amssymb}
\usepackage[ruled,linesnumbered]{algorithm2e}

\usepackage{microtype}

\aclfinalcopy 
\def\aclpaperid{426} 

\title{{A}dversarial {S}elf-Supervised Data-Free Distillation for Text Classification}

\author{Xinyin Ma, Yongliang Shen, Gongfan Fang, Chen Chen, Chenghao Jia, Weiming Lu\thanks{* Corresponding author} \\
College of Computer Science, Zhejiang University \\
\texttt{\{maxinyin, syl, fgf, chen\_double, chjia, luwm\}@zju.edu.cn} \\}

\date{}

\begin{document}
\maketitle
\begin{abstract}
Large pre-trained transformer-based language models have achieved impressive results on a wide range of NLP tasks. In the past few years, Knowledge Distillation(KD) has become a popular paradigm to compress a computationally expensive model to a resource-efficient lightweight model. However, most KD algorithms, especially in NLP, rely on the accessibility of the original training dataset, which may be unavailable due to privacy issues. 
To tackle this problem, we propose a novel two-stage data-free distillation method, named \textit{\textbf{Adversarial self-Supervised Data-Free Distillation}} (AS-DFD), which is designed for compressing large-scale transformer-based models (e.g., BERT). To avoid text generation in discrete space, we introduce a \textit{\textbf{Plug \& Play Embedding Guessing}} method to craft pseudo embeddings from the teacher's hidden knowledge. Meanwhile, with a \textit{\textbf{self-supervised module}} to quantify the student's ability, we adapt the difficulty of pseudo embeddings in an adversarial training manner. 
To the best of our knowledge, our framework is the first data-free distillation framework designed for NLP tasks. We verify the effectiveness of our method on several text classification datasets.

\end{abstract}

\section{Introduction}
Recently, pre-trained language models \citep{devlin2018bert,yang2019xlnet,liu2019roberta, raffel2019exploring} have achieved tremendous progress and reached the state-of-the-art performance in various downstream tasks such as text classification \citep{maas-EtAl:2011:ACL-HLT2011}, language inference \citep{bowman2015large} and question answering \citep{Rajpurkar2016SQuAD10}. These models become an indispensable part of current models for their transferability and generalizability.

However, such language models are huge in volume and demand highly in computational resources, making it impractical in deploying them on portable systems with limited resources (e.g., mobile phones, edge devices) without appropriate compression. Recent researches \citep{mccarley2019pruning, gordon2020compressing,michel2019sixteen} focus on compressing the large-scale models to a shallow and resource-efficient network via weight pruning \citep{guo2019reweighted}, knowledge distillation \citep{mukherjee2019distilling}, weight quantization \citep{zafrir2019q8bert} and parameter sharing \citep{Lan2020ALBERT:}.
Among them, some methods \citep{sanh2019distilbert,sun-etal-2019-patient} draw on the idea of transfer learning, utilizing knowledge distillation \citep{44873} to transfer latent representation information embedded in teachers to students. These knowledge distillation methods share some commonalities: they rely on the training data to achieve high accuracy. It will be intractable if we need to compress a model without publicly accessible data. Reasons for that include privacy protection, company assets, safety/security concerns and transmission. Representative samples include GPT2 \citep{radford2019language}, which has not released its training data with fears of abuse of language models. Google trains a neural machine translation system \citep{Wu2016GooglesNM} using internal datasets owned and protected by the company. DeepFace \citep{Taigman2014DeepFaceCT} is trained on user images under confidential policies for protecting users. Further, some datasets, like Common Crawl dataset used in GPT3 \citep{brown2020language}, contain nearly a trillion words and are difficult to transmit and store.

Conventional knowledge distillation methods are highly dependent on data.
Some models or algorithms in Computer Vision like DAFL \citep{chen2019data}, ZSKD \citep{nayak2019zero} solve the data-free distillation by generating pseudo images or utilizing metadata from teacher models. Exploratory researches \citep{micaelli2019zero, fang2019data} also show that GANs can synthesize harder and more diversified images by exploiting disagreements between teachers and students. However, these models only make attempts in image tasks, designing for continuous and real-valued images. Applying these models to generate sentences is challenging due to the discrete representation of words \citep{huszar2015not}. Backpropagation on discrete words is not reasonable, and it seems unlikely to pass the gradient through the text to the generator.
Apart from the discontinuity problem of text, some promotion strategies like layer-wise statistic matching in batch normalization \cite{yin2019dreaming} are not suitable for transformer-based models, which transposes batch normalization into layer normalization to fit with varied sentence length \citep{ba2016layer}.

To address the above issues and distill without data, we propose a novel data-free distillation framework called "Adversarial self-Supervised Data-Free Distillation"(AS-DFD). We invert BERT to perform gradient updates on embeddings and consider parameters of the embedding layer as accessible knowledge for student models. Under constraints of constructing  "BERT-like" vectors, pseudo embeddings extract underlying representations of each category. Besides, we employ a self-supervised module to quantify the student's ability and adversarially adjust the difficulty of pseudo samples, alleviating the insufficient supervisory problem controlled by the one-hot target. Our main contributions are summarized as follows:
\begin{itemize}
\item We introduce AS-DFD, a data-free distillation framework, to compress BERT. To the best of our knowledge, AS-DFD is the first model in NLP to distill knowledge without data.
\item We propose a Plug \& Play Embedding Guessing method and align the pseudo embeddings with the distribution of BERT's embedding. We also propose a novel adversarial self-supervised module to search for samples students perform poorly on, which also encourages diversity.
\item We verify the effectiveness of AS-DFD on three popular text classification datasets with two different student architectures. Extensive experiments support the conjecture that synthetic embeddings are effective for data-free distillation.
\end{itemize}

\section{Related Work}
\subsection{Data-Driven Distillation for BERT}
Knowledge Distillation (KD) compresses a large model (the teacher model) to a shallow model (the student model) by imitating the teacher's class distribution output ~\citep{44873}. Bert \citep{devlin2018bert} contains multiple layers of transformer blocks \citep{vaswani2017attention} which encodes contextual relationship between words. Recently, many works successfully compress BERT to a BERT-like model with knowledge distillation
~\citep{sanh2019distilbert} and achieve comparable performances on downstream-tasks.
Patient-KD~\citep{sun-etal-2019-patient} bridges the student and teacher model between its intermediate outputs.
TinyBERT~\citep{jiao2019tinybert} captures both domain-general and domain-specific knowledge in a two-stage framework.
\citet{zhao2019extreme} employs a dual-training mechanism and shared projection matrices to compress the model by more than 60x. 
BERT-of-Theseus~\citep{xu2020bert} progressively module replacing and involves a replacement scheduler in the distillation process. 
Besides, some recent surveys focus on compress BERT to a CNN-based \citep{chia2019transformer} or LSTM-based model to create a more lightweight model with additional training data \citep{tang-etal-2019-natural, tang2019distilling}.

\subsection{Data-Free Distillation Methods}
Current methods for data-free knowledge distillation are applied in the field of computer vision. 
~\citet{lopes2017data} leverages metadata of networks to reconstruct the original dataset.
~\citet{chen2019data} trains a generator to synthesize images that are compatible with the teacher. 
~\citet{nayak2019zero} models the output distribution space as a Dirichlet distribution and updated the random noisy images to compose a transfer set. 
~\citet{micaelli2019zero} and ~\citet{fang2019data} incorporate the idea of adversarial training into knowledge distillation, measuring the discrepancy between the student and teacher.
~\citet{yin2019dreaming} introduces DeepInversion to synthesize class-conditional images.
Due to the discrete nature of language, none of the above methods can be applied to natural language tasks. ~\citet{melas2020generation} proposes a generation-distillation framework in low-data settings, which employs a finetuned GPT2 as the generator and a CNN as the student model.
Different from methods above, we investigate the problem of compressing BERT with no data.

\begin{figure*}[h]
  \centering
  \includegraphics[scale=0.47]{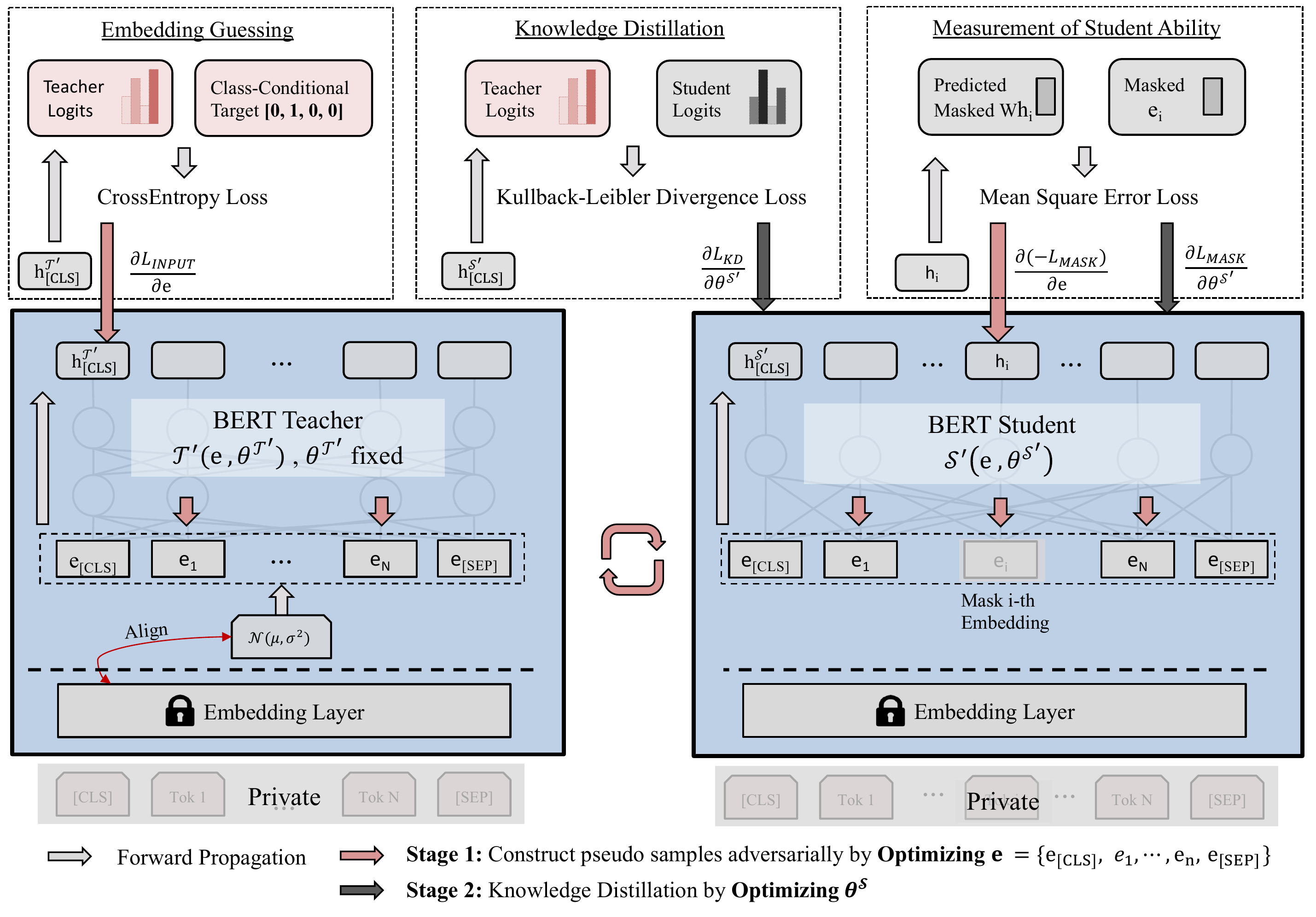}
  \caption{\label{model-graph} An overview of our two-stage Adversarial self-Supervised Data-Free Distillation framework. $\mathcal{T}^{\prime}$ and $\mathcal{S}^{\prime}$ contain transformer layers and classifier head. Firstly, when constructing synthetic samples, we iteratively guess and update the pseudo embeddings $\mathbf{e}$ under the feedback of the teacher's class-conditional supervision (top left) and the student's self-assessment (top right) in an adversarial training manner. Secondly, we use the generated sample $\mathbf{e}$ to distill knowledge (top middle). The parameters of embedding layer are fixed, and no inputs will go through the embedding layer when training.}
\end{figure*}

\section{Methods}
In this section, we present our two-stage distillation framework named \emph{Adversarial self-Supervised Data-Free Distillation} (AS-DFD).
We craft well-trained embedding-level pseudo samples by controllable Plug \& Play Embedding Guessing with alignment constraints (Section \ref{construct-set}) and adversarially adapt synthetic embeddings under self-supervision of the student (Section ~\ref{adversarial-student}). Using these pseudo samples, we transfer knowledge from the teacher to the student (Section ~\ref{KD}). The workflow of AS-DFD is illustrated in Figure~\ref{model-graph}.

\paragraph{Problem Definition}
Knowledge Distillation is a compression technique to train a high-performance model with fewer parameters instructed by the teacher model \citep{44873}.
Let $\mathcal{T}$ be a large transformer-based teacher model (12-layer BERT-base here) and $\mathcal{S}$ be a comparatively lightweight student model.
For each sentence $\mathbf{x}$, the classification prediction can be formulated as:
\begin{equation}
  \begin{array}{l}
    \mathbf{e} = \operatorname{EmbeddingLayer}(\mathbf{x};\hspace{0.5 mm} \theta_{emb}) \\
    \mathbf{h} = \operatorname{TransformerLayers}(\mathbf{e}; \hspace{0.5 mm}\theta_{layer}) \\ 
    \mathbf{y} = \operatorname{ClassifierLayer}(\mathbf{h_{[CLS]}} ; \hspace{0.5 mm}\theta_{classifier})
  \end{array}
\end{equation}
where $\theta_{emb}$, $\theta_{layer}$, $\theta_{classifier}$ represent parameters in the embedding layer, transformer layers and classification head respectively. $\mathbf{y}$ is the softmax probability output of $\mathbf{x}$ and $\mathbf{h_{[CLS]}}$ denotes the hidden states in the last layer corresponding to the special token [CLS].
Parameters with superscript $\mathcal{T}$ belong to the teacher and $\mathcal{S}$ for the student.

Our goal of data-free knowledge distillation is to train the student parameters $\theta^{\mathcal{S}}$ with no data $\mathcal{X}$ available. In other words, we only have a teacher model $\mathcal{T}$ and we need to compress it.

\subsection{Construct Pseudo Samples}
\label{construct-set}
\paragraph{Plug \& Play Embedding Guessing}
In the data-free settings, we need to solve the dilemma of having no access to the original dataset. The major challenge is how to construct a set of highly reliable samples, from which the student can extract differential knowledge.

Our approach exploits representative knowledge hidden in the teacher's parameters in a Plug \& Play manner \citep{nguyen2017plug, Dathathri2020plug}. 
Given a sentence $\textbf{x}$ and a label $\textbf{y}$, the conditional probability can be written as $P(\textbf{y}|\textbf{x}; \theta^{\mathcal{T}})$. When finetuning the teacher, we optimize parameters $\theta^{\mathcal{T}}$ towards higher probability. To capture impression of prior training data in the teacher's parameters, we invert the model and utilize the teacher's parameters to guide the generation of $\mathbf{x}$ by ascending $P(\textbf{y}|\textbf{x}; \theta^{\mathcal{T}})$ with $\theta^{\mathcal{T}}$ fixed. 

Due to the intractable discrete problem of text, gradients updated on $\mathbf{x}$ are pointless. Most language models transform discrete words into continuous embeddings. Inspired by this, we ignore the embedding layer and apply the updating on continuous representation space of embeddings. We name this generation process "Embedding Guessing". We randomly guess vectors $\mathbf{e} \in \mathbb{R}^{l \times d}$,  feed them into the transformer blocks and get feedback from gradients to confirm or update our guess. $l$ is the predefined length of sentence and $d$ is the embedding dimensionality, which is 768 in BERT-base. Those target-aware embeddings can be obtained by minimizing the objective:
\begin{equation}
    \mathcal{L}_{INPUT} = \sum_{\mathbf{e} \in \mathbf{\mathcal{E}}} \operatorname{CE}\left(\mathcal{T^\prime}(\mathbf{e}; \theta^{\mathcal{T}^{\prime}}), \hat{\mathbf{y}}\right)\\
\end{equation}
where $\mathcal{T}^{\prime}$ takes pseudo embeddings $\mathbf{e}$ as input and contains $\operatorname{TransformerLayers}$ and $\operatorname{ClassifierLayer}$ in the teacher. $\theta^{\mathcal{T}^{\prime}}$ includes $\theta_{layer}^{\mathcal{T}}$ and $\theta_{classifier}^{\mathcal{T}}$.
$\mathbf{\hat{y}}$ is a random target class. $\operatorname{CE}$ refers to the cross-entropy loss. $\mathcal{E}$ is a batch of $\mathbf{e}$ initialized with Gaussian distribution. We update $\mathbf{e}$ for several iterations until convergence, representing that $\mathbf{e}$ is correct judged by the teacher. As for $\theta^\mathcal{S}_{emb}$, we share $\theta^\mathcal{T}_{emb}$ with $\theta^\mathcal{S}_{emb}$.

We argue that under the process of Embedding Guessing, pseudo embeddings $\mathbf{e}$ contain the target-specific information.
Classification models need to find out differentiated characteristics which propitious to prediction over multiple categories.
As the human learning process, examples given by teachers are encouraged to be representative and better reflecting the discrepancy among classes. Borrowed from this teaching strategy, we guess embeddings towards the direction of higher likelihood on target category and seek the local minimum regarding the target class, which reflects the characteristics of the target class within regions. In other words, these synthetic samples are more likely to comprise separation statistics between classes.

\paragraph{Making Pseudo-Embeddings More Realistic}
However, training on embeddings leads to a gap between the pseudo embeddings and the true underlying embeddings. Specifically, Embedding Guessing is independent of the parameter of the teacher's embedding and will shift the representational space. We add some additional constraints to ensure generated embeddings imitate the distribution of real data to a certain extent. Alignment strategies to restrain and reduce search space are listed as follows:
\begin{itemize}
  \item Add $\mathbf{e_{[CLS]}}$ and $\mathbf{e_{[SEP]}}$ at both ends of the synthetic embeddings. $\mathbf{e_{[CLS]}}$ and $\mathbf{e_{[SEP]}}$ represent embeddings corresponding to [CLS] and [SEP].
  \item Continuously mask random length of embeddings from the tail of it. Lengths of sentences in batches are indeterminate and synthetic embeddings should cover this scenario.
  \item Adjust the Gaussian distribution to find the best initialization. Excessive initialization scope expands search space while small one converges to limited samples.
\end{itemize}

\subsection{Adversarial self-Supervised Student}
\label{adversarial-student}
\paragraph{Modeling Learning Ability of the Student}
Effective teaching needs to grasp the student's current state of knowledge and dynamically adapt teaching strategies and contents. How to model the ability of the student without data? While processing natural language, the ability to analyze the context is an indicator of the student's capabilities and it can be quantified by a self-supervised module. Borrowing the idea of masking and predicting the entries randomly, we randomly mask one embedding in $\mathbf{e}$. Then, a new self-restricted objective is to predict the masked embedding with the following forums:

\begin{equation}
  \begin{array}{l}
    \mathbf{h} = \mathcal{S^\prime}\left(\mathbf{e}^{mask}; \theta^{\mathcal{S}^{\prime}}\right) \\

    L_{MASK} = \displaystyle \sum\limits_{\mathbf{e} \in \mathcal{E}}\left\|\frac{\mathbf{e}_i}{\left\|\mathbf{e}_i\right\|_{2}}-
                                            \frac{W\mathbf{h}_i}{\left\|W\mathbf{h}_i\right\|_{2}}\right\|_{2}^{2} \\
  \end{array}
\end{equation}
where $\mathbf{e}$ is randomly masked on position $i$ and converted to $\mathbf{e}^{mask}$. $\mathbf{e}_i$  is the masked embedding and $W$ is the parameters in the fully-connected supervised module for predicting masked embedding. $\mathcal{S}^{\prime}$ acts the same way as $\mathcal{T}^{\prime}$. Unlike the class-conditional guidance, the self-supervised module shifts the gradients with more concrete and diverse supervision from context.

\paragraph{Adversarial Training of the Student}
To enforce $\mathbf{e}$ with more valuable and diverse information, we encourage the student to adversarially search for samples that the student is not confident. Prior works \citep{micaelli2019zero,fang2019data} maximize the discrepancy between the teacher and student to encourage difficulty in samples and avoid synthesizing redundant images. We design a self-assessed confrontational mechanism, which guides the pseudo embeddings towards greater difficulty by enlarging $L_{MASK}$ in the constructing stage and enhances the student by decreasing $L_{MASK}$ in the distillation stage. Here, $L_{MASK}$ acts as the timely student's feedback to improve teaching.

\subsection{Two-stage Training}
\label{KD}
\paragraph{Distillation Objective}
Students learn high-entropy knowledge from teachers by matching soft targets.
Taking $\mathbf{\mathcal{E}}$ as synthetic samples, we measure the distance between the teacher and student as:
\begin{equation}
  L_{KL} = \sum_{\mathbf{e} \in \mathbf{\mathcal{E}}} \operatorname{KL}
  \left(
     \mathcal{T^\prime}(\mathbf{e}; \theta^{\mathcal{T}^{\prime}}), 
     \mathcal{S^\prime}(\mathbf{e}; \theta^{\mathcal{S}^{\prime}}), \tau
  \right)
\end{equation}
where $\operatorname{KL}$ denotes the Kullback-Leibler divergence loss and $\tau$ is the distillation temperature.

We follow PKD \citep{sun-etal-2019-patient} to learn more meticulous details for students. To capture rich features, we define the additional loss as:
\begin{equation}
 L_{PT} = \sum_{\mathbf{e} \in \mathbf{\mathcal{E}}}\left\|
                \frac{{\mathbf{h_{[CLS]}}}^{\mathcal{T}}}{\left\|{\mathbf{h_{[CLS]}}}^{\mathcal{T}}\right\|_{2}}-
                \frac{{\mathbf{h_{[CLS]}}}^{\mathcal{S}}}{\left\|{\mathbf{h_{[CLS]}}}^{\mathcal{S}}\right\|_{2}}\right\|_{2}^{2}
\end{equation}
The objective of distillation can be formulated as:
\begin{equation}
  L_{KD} = L_{KL} + \alpha L_{PT}
\end{equation}
where $\alpha$ balances these two losses.

\paragraph{Training Procedure}
We summarize the training procedure in algorithm~\ref{algorithm}. The multi-round training of AS-DFD splits into two steps: the construction stage and the distillation stage.
In the construction stage, after randomly sampling vectors with alignment constraints, we repeat the adversarial training of pseudo embeddings for $n_{iter}$ times. In each iteration, we guess embeddings under class-conditional supervision information for $n_{\mathcal{T}}$ steps, and the student is asked to predict and give negative feedback to guide pseudo-embeddings' generation for $n_{\mathcal{S}}$ steps. 
When distilling, we train $\theta^{\mathcal{S}^{\prime}}$ as well as $W$ with those pseudo samples.

\begin{algorithm}[h]
  \caption{\label{algorithm} Two-stage Adversarial self-Supervised Data-Free Distillation}
  \SetAlgoLined
  \KwIn{Teacher model $\mathcal{T}$ with $\theta^{\mathcal{T}}$, $\mu$, $\sigma$}
  \KwOut{Student model $\mathcal{S}$ with $\theta^{\mathcal{S}}$, $W$}
  \BlankLine
  Initial $\theta^{\mathcal{S}^{\prime}}$ with $\theta^{\mathcal{T}^{\prime}}$ and set $\theta^{\mathcal{S}}_{emb} \leftarrow \theta^{\mathcal{T}}_{emb}$ \\
  \For{$i\leftarrow 1$ \KwTo $N$}{
    \emph{// Stage 1: Construct Pseudo Samples} \\
    Fix $\theta^{\mathcal{T}^{\prime}}$, $\theta^{\mathcal{S}^{\prime}}$ and $W$ \\
    Sample $\mathcal{E} \sim \mathcal{N}(\mu, \sigma^{2})$\\
    Add alignment constraints on $\mathcal{E}$  \\
    \For{iters $\leftarrow 1$ \KwTo $n_{iter}$}{
      \For{$m \leftarrow 1$ \KwTo $n_\mathcal{T}$}{
            $\mathcal{E} \leftarrow \mathcal{E} - \eta \displaystyle \frac{\partial L_{INPUT}}{\partial \mathcal{E}} $
      }
      \For{$n \leftarrow 1$ \KwTo $n_\mathcal{S}$}{
        $\mathcal{E} \leftarrow \mathcal{E} - \eta \displaystyle \frac{\partial \left(-L_{MASK}\right)}{\partial \mathcal{E}}$ 
      }
    }
    \BlankLine
    \emph{// Stage 2: Knowledge Distillation} \\
    Fix $\theta^{\mathcal{T}^{\prime}}$ and update $\theta^{\mathcal{S}^{\prime}}$, $W$ \\
    
    $\theta^{\mathcal{S}^{\prime}} \leftarrow \theta^{\mathcal{S}^{\prime}} - \xi \displaystyle \frac{\partial L_{KD}}{\partial \theta^{\mathcal{S}^{\prime}}} $ \\
    $W \leftarrow W - \xi \displaystyle \frac{\partial L_{MASK}}{\partial W} $
  }
\end{algorithm}

\section{Experiments}

\subsection{Datasets}
We demonstrate the effectiveness of our methods on three widely-used text classification datasets:
AG News, DBPedia, IMDb \citep{auer2007dbpedia,maas-EtAl:2011:ACL-HLT2011}. The statistics of these datasets are shown in Table~\ref{dataset-statistic}.
For datasets without validation sets (DBPedia and IMDb), we randomly sample 10\% of the train set as the validation set.

\begin{table}[h]
    \centering
    \begin{tabular}{ccccc}
    \hline
    Dataset  & Classes & Train & Valid &Test\\
    \hline
    AG News  &  4      & 114k  & 6k   & 7.6k \\
    DBPedia  &  14     & 504k  & 56k  & 70k\\ 
    IMDb     &  2      & 22.5k & 2.5k &25k\\
    \hline
    \end{tabular}
    \caption{\label{dataset-statistic} Statistics of AG News/DBPedia/IMDb. Training samples are only available when finetuning teacher models. AG News and DBPedia are topic classification datasets and IMDb is a dataset for binary sentiment classification.}
\end{table}

\begin{table*}
  \centering
  \begin{tabular}{lccc}
  \hline
  \hline
  &AG News&DBPedia&IMDb\\
  \hline
  \hline
  \emph{Distill on Original Dataset} \\
  Teacher - $\text{BERT}_{12}$ & 94.2 & 99.4  & 88.5\\
  Student - $\text{BERT}_{6}$  & 94.1 & 99.3  & 87.0\\
  Student - $\text{BERT}_{4}$  & 93.8 & 99.3  & 85.9\\
  \hline
  \emph{Data-Free Distillation - $\text{BERT}_6$ as student} \\
  Random Text                  &85.4      & 93.9  & 77.1 \\
  Modified-ZSKT                &88.4      & -  & 78.1 \\
  Modified-ZSKD                &88.6      & 97.1  & 78.2 \\
  AS-DFD (Ours)                &\textbf{90.4}     & \textbf{98.2}  & \textbf{79.8} \\
  \hline
  \emph{Data-Free Distillation - $\text{BERT}_4$ as student} \\
  Random Text                   &78.5     & 77.3  & 67.6 \\
  Modified-ZSKT                 &81.1     & -     & 70.4 \\
  Modified-ZSKD                 &83.8     & 83.0  & 70.7 \\
  AS-DFD (Ours)                 &\textbf{88.2}    & \textbf{94.1}  & \textbf{77.2}\\
  \hline
  \hline
  \end{tabular}
  \caption{\label{accuracy}Distillation accuracy on three datasets: AG news, DBPedia and IMDb.
 For fair comparision, Modified-ZSKT and Modified-ZSKD synthetic embeddings rather than images compared with its original algorithm. '-' means that accuracy cannot exceed the result of Random Text. Results show that AS-DFD outperforms other baselines in data-free distillation.}
\end{table*}

\subsection{Teacher/Student Models}
We experiment with official uncased BERT-base \citep{devlin2018bert} as the teacher model ($\text{BERT}_{12}$) for its widespread use in downstream tasks. 
BERT-base has 12 layers of Transformer \citep{vaswani2017attention} with 12 attention heads in each layer. 
We conduct experiments on student models with different transformer layers: 4-layer BERT ($\text{BERT}_4$) or 6-layer BERT ($\text{BERT}_6$). Statistics of parameters and inference time are listed in Table ~\ref{model-statistic}.
\begin{table}[h]
    \centering
    \begin{tabular}{ccc}
    \hline
    Layers & Params & Inference Time(s)\\
    \hline
    12 & 109M (1$\times$)     &  26.9s (1$\times$)\\
    6  &  67M (1.63$\times$)  &  14.1s (1.91$\times$)\\ 
    4  &  52M (2.10$\times$)  &  9.5s  (2.84$\times$)\\
    \hline
    \end{tabular}
    \caption{\label{model-statistic} Number of parameters and inference time for $\text{BERT}_{12}$, $\text{BERT}_{6}$ and $\text{BERT}_{4}$. Inference speed is tested on 7.6K samples from AG News.}
\end{table}

\begin{figure*}[h]
  \centering
  \includegraphics[scale=0.40]{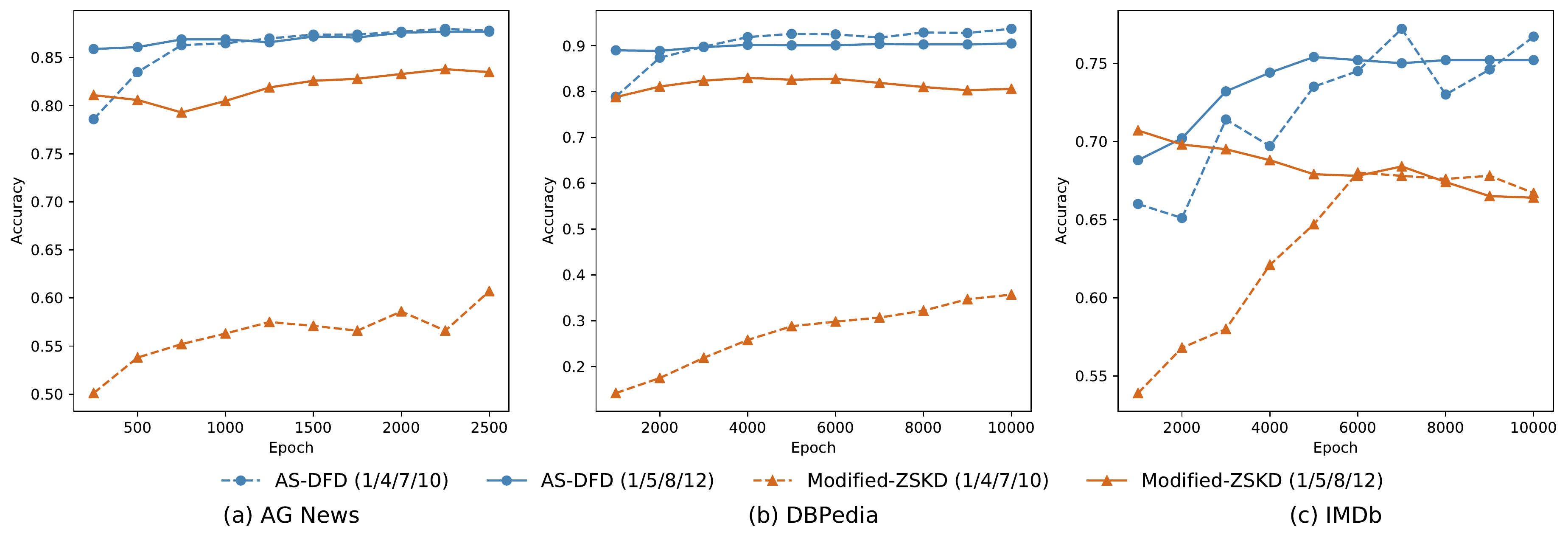}
  \caption{\label{initialization}Comparison of AS-DFD versus Modified-ZSKD using different initializations. Experiments are conducted on the 4-layer BERT student. 
           Dash lines show the result initialized with BERT's \{1, 4, 7, 10\} layers and solid lines with \{1, 5, 8, 12\} layers. }
\end{figure*}

\subsection{Baselines}
To the best of our knowledge, there is no data-free distillation method for language tasks. However, when slightly modifying the data-free distillation models that are effective in Computer Vision, these models can also work on language tasks. Imitating Plug \& Play Embedding Guessing method, we plug those image generators/generation methods above the embedding layer to synthesize continuous embeddings (instead of images).

Except for a baseline of random selection of words, we choose two models that represent the mainstream approaches in data-free distillation of image classification. Baselines are described as follows:

\paragraph{Random Text}We randomly select words from vocabulary and construct literally-uninterpretable sentences. 
\paragraph{Modified-ZSKT}Modified-ZSKT is extended from ZSKT \citep{micaelli2019zero}. ZSKT trains an adversarial generator to search for images in which the student's prediction poorly matches that of the teacher's and reaches state-of-the-art performance. 
\paragraph{Modified-ZSKD} Modified-ZSKD is derived from ZSKD \citep{nayak2019zero}. ZSKD performs Dirichlet sampling on class probability and craft Data Impression. DeepInversion \citep{yin2019dreaming} extends ZSKD with feature distribution regularization in batch normalization and outperforms ZSKD. However, BERT is not suitable for this performance-enhancing approach (BERT has no BN or structure like BN to store statistics of training data) and DeepInversion cannot be the baseline of our method.

\subsection{Experimental Results}
We first show the performance of data-driven knowledge distillation. Then we show the effectiveness of AS-DFD methods. 
As shown in Table~\ref{accuracy}, AS-DFD with $\text{BERT}_4$ and $\text{BERT}_6$ performs the best on three datasets. For 6-layer BERT, our algorithm improves 1.8\%, 1.1\% and 1.6\% compared to Modified-ZSKD, closing the distance between the teacher and student.
Furthermore, when coaching the 4-layer student, our methods gain 4.4\%, 11.1\% and 6.5\% increases, which significantly improves the distillation accuracy. It seems that AS-DFD performs better with higher compression rates compared with other data-free methods. However, there is still a large gap between the performance of data-drive distillation and data-free distillation. 

As for other baselines, Random Text can be regarded as a special case of unlabeled text where models can extract information to infer on, especially on text classification tasks. We use it as a criterion to judge whether a model works.
Modified-ZSKT performs worse than Random Text on DBPedia. The reason lies in the structure of the generator, which is designed for image generation and is not suitable for language generation. The strength of CNN-based generators lies in its ability to capture local and hierarchical features. However, it is difficult for CNN to capture global and sequential structures, which is essential for languages.

\paragraph{Implementation Details}
We train the AS-DFD with $n_{\mathcal{T}} = 5, n_{\mathcal{S}} = 1$ and $n_{iter} = 5$. Maximum sequence lengths for three datasets are set to 128. 
Ideally, the more samples generated, the higher the accuracy. We impose restrictions on the number of generated samples for each dataset. Training epochs are 2.5k(AG News), 10k(DBPedia), 10k(IMDb) with 48 samples per batch for all methods except ZSKT, which needs to train its generator from scratch (25k epochs in Modified-ZSKT). In our experiments, these samples are enough for models to reach a stable status. More implementation details about finetuning teachers and distilling students are listed in Appendix~\ref{hyperparameters}.

\paragraph{Initialization}
We observe that students' performance is highly sensitive to initializations (especially the Random Text baseline). \citet{fan2019reducing} argues that different layers play different roles in BERT.
We report results using different initialization schemes and show the stability of AS-DFD.
Considering that the embedding layer is separated from transformer blocks when training, we strongly recommend sharing the first layer's parameters of the teacher with the student, which is also suggested in \citet{xu2020bert}.
Specifically, we choose two sets of layer weights. One is \{1, 4, 7, 10\}, which is common in data-driven distillation, and the other is \{1, 5, 8, 12\}, which intentionally put the last layer's parameters in.
We evaluate these initialization schemes on AS-DFD and Modified-ZSKD. To eliminate the effects of distillation, we ensure that hyperparameters in the distillation step are consistent in two models, which intuitively shows the disparity in samples' quality.
We do not include Modified-ZSKT because samples of Modified-ZSKT vastly outnumber the other two approaches.

Experimental results are shown in Figure~\ref{initialization}. Modified-ZSKD highly dependent on initialization, especially on AG news and DBPedia with 23.1\% and 47.1\% performance drop relatively. On the contrary, initialization has limited impacts on AS-DFD. If pseudo-embeddings are initialized with worse parameters, our method still achieves better accuracy than other baselines (87.7\% on AG News, 90.5\% on DBPedia and 75.4\% on IMDb). It shows that our method synthesizes higher-quality samples compared with Modified-ZSKD. Additionally, AS-DFD maintains an upward trend when the size of synthetic samples grows, suggesting that synthetic samples are useful for knowledge transfer.

\paragraph{Validity of Synthetic Embeddings}
\begin{figure}[t]
  \centering
  \includegraphics[scale=0.56]{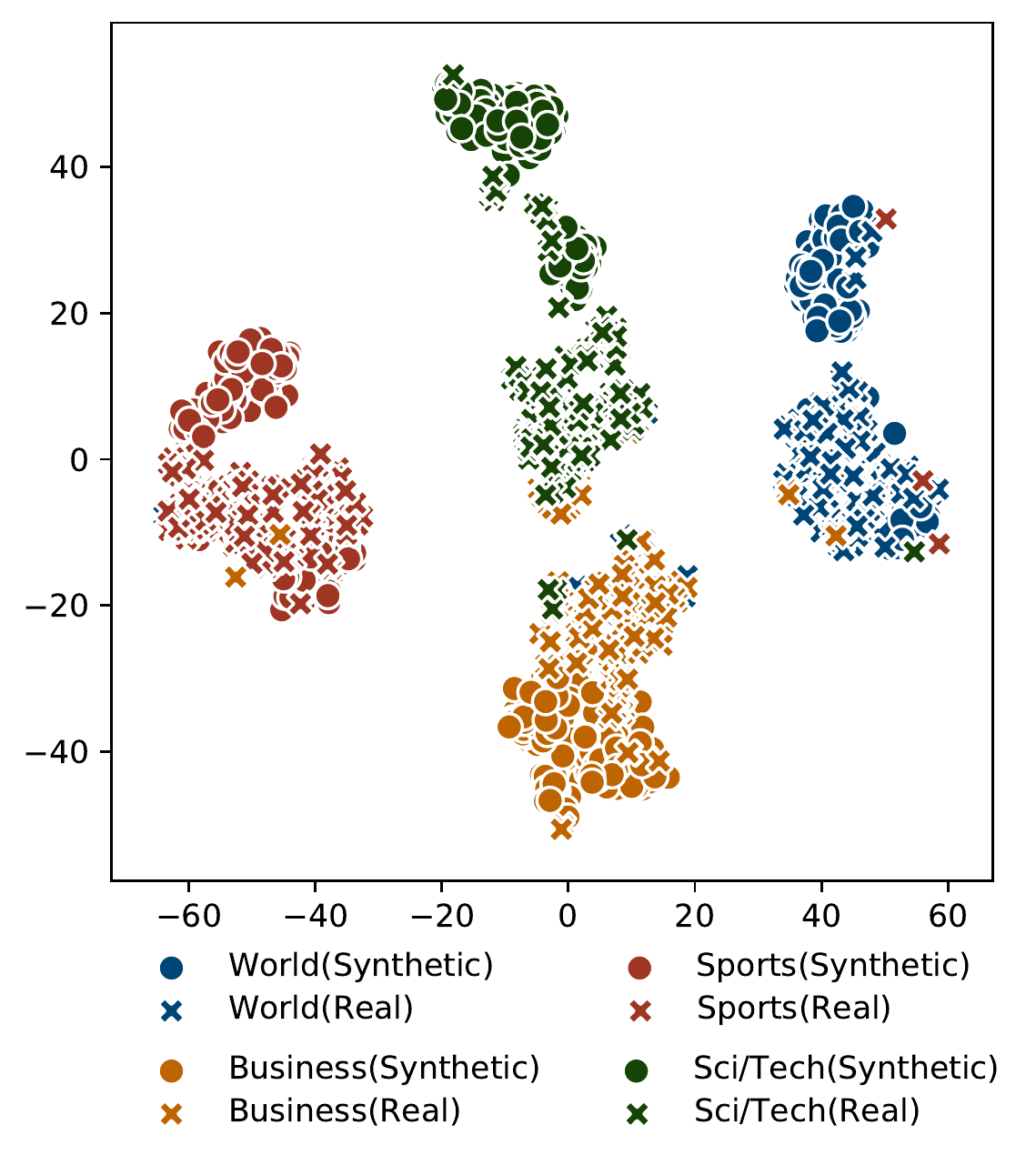}
  \caption{\label{t-SNE-validity}t-SNE dimensionality reduction results between real and synthetic samples on output of last layer. }
\end{figure}
Embeddings we generated are incomprehensible. We use t-SNE \citep{maaten2008visualizing} to visualize the synthetic embeddings in comparison with the original dataset. As shown in Figure~\ref{t-SNE-validity}, samples generated by Embedding Guessing are close to the real samples and overlap with them to a certain extent.

\subsection{Module Analysis}
To verify the contribution of each module, we perform an ablation study and summarize it in Table~\ref{ablation}.

Embedding Guessing is the foundation of the entire model. After drawing into the idea of Plug \& Play Embedding Guessing, distillation performance is improved with stability, demonstrating that knowledge extracted from the teacher makes the synthetic samples reasonable. The embedding layer of the student model is completely separated in the generation-distillation process. Imitating BERT's input precisely narrows this gap, leading to a large improvement in accuracy. 
Additionally, choosing an appropriate normal distribution can effectively reduce search space and avoid generating completely irrelevant samples. We conduct experiments on different normal distributions in Appendix~\ref{gaussian}.

\begin{table}[t]
  \centering
  \begin{tabular}{lc}
  \hline
  \hline
  Method & Accuracy \\ 
  \hline
  \hline
  Random Noise &  25.1\\
  \hline
  + Embedding Guessing & 44.2\\
  \hline
  + Alignment Constraints & \\
    \hspace{1.5 mm}  + Add [CLS] and [SEP] & 80.3\\
    \hspace{1.5 mm}  + Variable Length & 82.2 \\
    \hspace{1.5 mm}  + Appropriate Gaussian Distribution & 87.4\\
  \hline
  + Adversarial self-Supervised Module& 88.2\\
  \hline
  \hline
  \end{tabular}
  \caption{\label{ablation} Ablation study on AG News dataset. The Student model $\text{BERT}_4$ is initialized with BERT's 1st, 4th, 7th and 10th layers.}\label{tab:accents}
\end{table}

\begin{figure}[t]
  \centering
  \includegraphics[scale=0.56]{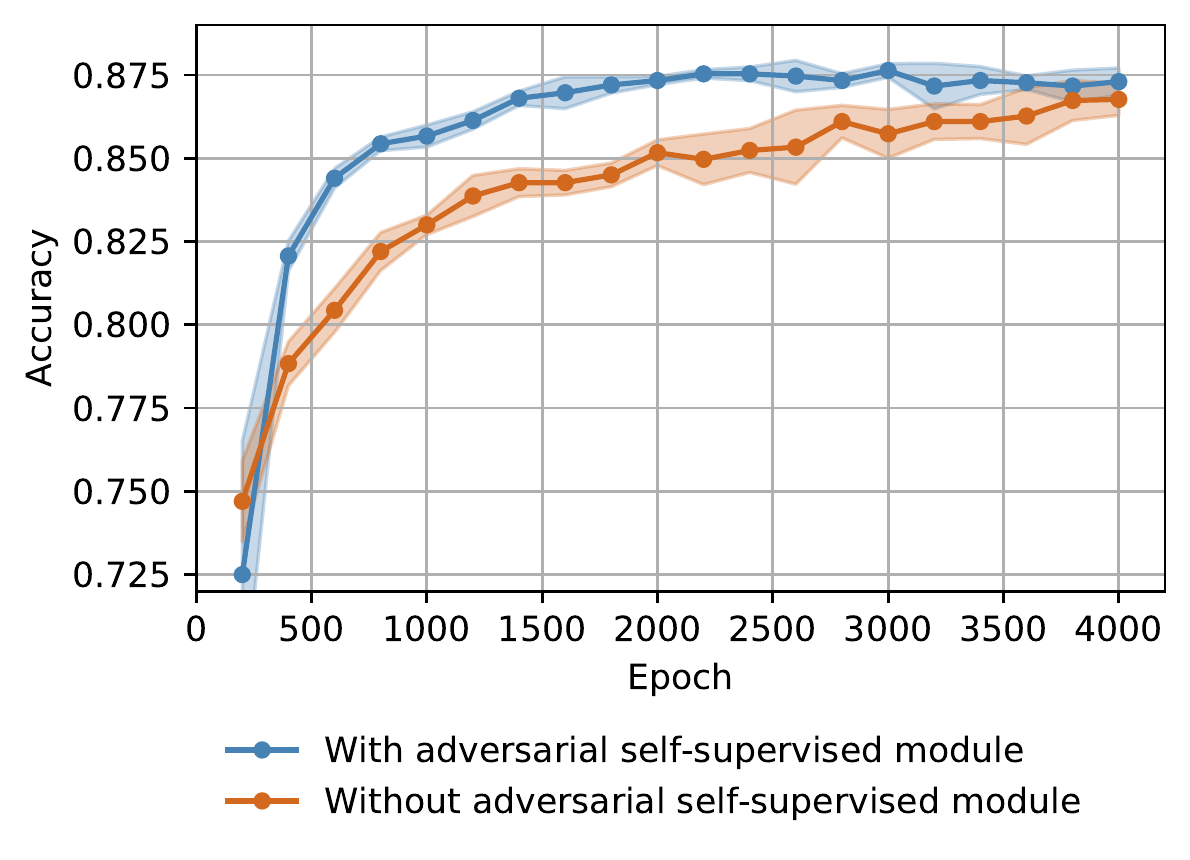}
  \caption{\label{adversarial} Accuracy curve with / without adversarial self-supervised module. The shaded area around the curve is the standard deviation over three seeds.}
\end{figure}

\paragraph{Effect of Adversarial self-Supervised Module}
To investigate whether the adversarial self-supervised module help data-free distillation, 
we conduct experiments on AG News to demonstrate the advantage of it in Figure~\ref{adversarial}.

We repeat each experiment 3 times and plot mean and standard deviation to reduce the contingency of experiments.
With the adversarial self-supervised module, distillation converges faster and achieves higher accuracy. The number of epochs can be reduced to 2500, saving half of the time.
As shown in the curve, AS-DFD does not perform well in the early stage since the self-supervised module is underfitting. After training for a while, the self-supervised module can grasp the student's ability and provide corrective feedback to synthesize more challenging samples.

\section{Conclusion}
In this paper, we propose AS-DFD, a novel data-free distillation method applied in text classification tasks. We use Plug \& Play Embedding Guessing with alignment constraints to solve the problem that gradients cannot update on the discrete text. To dynamically adjust synthetic samples according to students' situations, we involve an adversarial self-supervised module to quantify students' abilities. Experimental results on three text datasets demonstrate the effectiveness of AS-DFD.

However, it's still challenging to ensure the diversity of generated embeddings under the weak supervision signal and we argue that the gap between synthetic and real sentences still exists. 
In the future, we would like to explore data-free distillation on more complex tasks.

\section*{Acknowledgments}
This work is supported by the National Key Research and Development Project of China (No. 2018AAA0101900), the Fundamental Research Funds for the Central Universities (No. 2019FZA5013), the Zhejiang Provincial Natural Science Foundation of China (No. LY17F020015), the Chinese
Knowledge Center of Engineering Science and Technology (CKCEST) and MOE Engineering Research Center of Digital Library.

\bibliography{anthology,emnlp2020}
\bibliographystyle{acl_natbib}

\appendix

\section{Appendices}
\label{sec:appendix}

\begin{figure*}[b]
  \centering
  \includegraphics[scale=0.5]{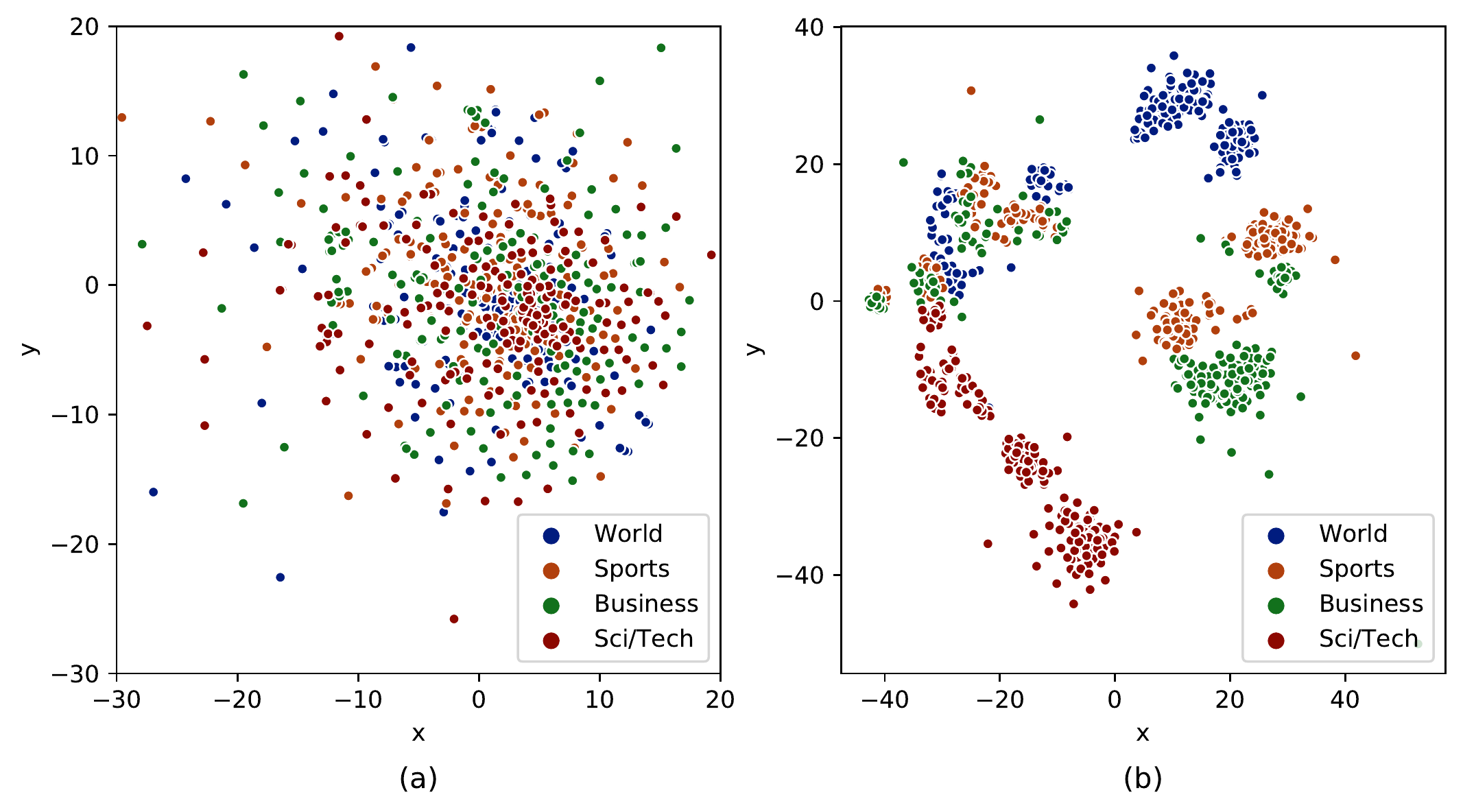}
  \caption{\label{gaussian-graph}t-SNE results on real samples(a) or synthetic samples(b)}
\end{figure*}

\subsection{Implementation Details}
\label{hyperparameters}
\paragraph{Hyperparameters in Finetuning Teachers}
We finetune BERT-base on three datasets mentioned above. 
We train our teacher models with Adam \citep{kingma2014adam} in 4 epochs.
Learning rate is set to 2e-5 with a scheduler that linearly decreases it after 10\% warmup steps. We set the maximum sequence length to 128 and batch size to 32 for all datasets.

\paragraph{Hyperparameters in Data-Free Distillation}
AS-DFD is trained on 1 TITAN Xp GPU. We set batch size to 48 with the student's learning rate $\xi$ from \{$5 \times 10^{-5}$, $2 \times 10^{-5}$, $1 \times 10^{-5}$\} and embedding learning rate $\eta$ from \{$1 \times 10^{-2}$, $5 \times 10^{-3}$, $1 \times 10^{-3}$\}. We conduct an additional search over $\alpha$ from \{100, 200, 250, 350, 500\} and select the hyperparameters with the highest accuracy. In our experiment, $\eta$ equals to $1 \times 10^{-2}$ and $\xi$ equals to $1 \times 10^{-5}$. $\alpha$ is set to 250. Temperature $\tau = 1$ works well in our model. In the distillation step, we use Adam with a warmup proportion of 0.1 and we linearly decay the learning rate. In the construction step, the learning rate is fixed with Adam optimizer. There may be no validation set under data-free settings, which makes tuning parameters impossible. We experiment with the hyperparameters performed best on AG News and find that this set of parameters also performs well on the other two datasets.

\subsection{Adjust Gaussian Distributions}
\label{gaussian}
The other two parameters are the mean and standard deviation for Gaussian sampling. We found in our experiments that standard deviation has a great influence on the student's performance.
If vectors are initialized with small standard deviation(e.g. std=0.05, see Figure~\ref{gaussian-graph}.b), generated samples in each category gather together, meaning that they aggregate to limited regions and leading to insufficient diversity of pseudo samples. Real data samples show no aggregation under t-SNE(see Figure~\ref{gaussian-graph}.a). A higher standard deviation(e.g. std=1) indicates that samples are spread out from the mean, which will increase the search space and far from the embedding's distribution of BERT. It is also reflected in our testing accuracy with 83.2, 85.3, 88.2, 83.2 corresponding to $\mathcal{N}(0, 0.05^{2})$, $\mathcal{N}(0, 0.2^{2})$, $\mathcal{N}(0, 0.35^{2})$, $\mathcal{N}(0, 1^{2})$.
We search standard deviations over \{0.05, 0.1, 0.2, 0.25, 0.3, 0.35, 0.4, 0.5, 1\} and choose 0.35 to be the best standard deviation, which works well on all three datasets.

\end{document}